\documentclass[runningheads]{llncs}
\usepackage{amsmath,amssymb}
\usepackage{graphicx}
\usepackage[misc]{ifsym}
\usepackage{times}
\usepackage{soul}
\usepackage{url}
\usepackage{bbm}
\begin{document}
%
%\title{Contribution Title\thanks{Supported by organization x.}}
\title{Understanding Difficulty-based Sample Weighting with a Universal Difficulty Measure\thanks{This study is supported by NSFC 62076178, TJF 19ZXAZNGX00050, and Zhijiang Fund 2019KB0AB03. \\ Paper published at ECML PKDD 2022}}
\author{Xiaoling Zhou\inst{1} \and
Ou Wu\Letter\inst{1} \and
Weiyao Zhu\inst{1}\and
Ziyang Liang\inst{1}
}
\authorrunning{Xiaoling Zhou et al.}
\institute{Center for Applied Mathematics, Tianjin University, China.
\email{\{xiaolingzhou,wuou\}@tju.edu.cn},\\
\email{weiyaozhu042@outlook.com},
\email{ziyangliang@tju.edu.cn}
}
\maketitle 
\begin{abstract}
Sample weighting is widely used in deep learning. A large number of weighting methods essentially utilize the learning difficulty of training samples to calculate their weights. In this study, this scheme is called difficulty-based weighting. Two important issues arise when explaining this scheme. First, a unified difficulty measure that can be theoretically guaranteed for training samples does not exist. The learning difficulties of the samples are determined by multiple factors including noise level, imbalance degree, margin, and uncertainty. Nevertheless, existing measures only consider a single factor or in part, but not in their entirety. Second, a comprehensive theoretical explanation is lacking with respect to demonstrating why difficulty-based weighting schemes are effective in deep learning. In this study, we theoretically prove that the generalization error of a sample can be used as 
a universal difficulty measure. %because it reflects most of the factors influencing the samples' learning difficulty.
Furthermore, we provide formal theoretical justifications on the role of difficulty-based weighting for deep learning, consequently revealing its positive influences on both the optimization dynamics and generalization performance of deep models, which is instructive to existing weighting schemes.

\end{abstract}
\section{Introduction}
Treating each training sample
unequally improves the learning performance. Two cues are typically considered in designing the weighting schemes of training samples~\cite{zhou2021samples}. The first cue is the application context of learning tasks. In applications such as medical diagnosis, samples with high gains/costs are assigned with high weights~\cite{SalmanHKhan44}. The second cue is the characteristics of the training data. For example, samples with low-confidence or noisy labels are assigned with low weights. Characteristic-aware weighting has attracted increasing attention owing to its effectiveness and universality~\cite{MPawanKumar04,TsungYiLin03,YoshuaBengio15}.

Many existing characteristic-aware weighting methods are based on an intrinsic property of the training samples, i.e., their learning difficulty. The measures for the samples' learning difficulty can be roughly divided into five categories.
%\vspace{-0.02in}
\begin{itemize}
    \item Prediction-based measures. This category directly uses the loss~\cite{MPawanKumar04,WenjieWang18,ThibaultCastells21} or the predicted probability of the ground truth~\cite{TsungYiLin03,EmanuelBenBaruch10} as the difficulty measures. This measure is simple yet effective and is widely used in various studies~\cite{MPawanKumar04,TsungYiLin03}. Their intention is that a large loss (a small probability) indicates a large learning difficulty. %Some methods use loss per epoch, while some others utilize the average loss during the training process.
    \item Gradient-based measures. This category applies the loss gradient in the measurement of the samples' learning difficulty~\cite{CarlosSantiagoa19,BuyuLi11}. Santiagoa et al. \cite{CarlosSantiagoa19} uses the norm of the loss gradient as the difficulty measure. Their intuition is that the larger the norm of the gradient, the harder the sample.
    \item Category proportion-based measures. This category is mainly utilized in imbalanced learning~\cite{YinCui17}, where the category proportion measures the samples' difficulty. People believe that the smaller the proportion of a category, the larger the learning difficulty of samples in this category~\cite{YinCui17,zhang2021distribution}.
    \item Margin-based measures. The term ``margin" refers to the distance from the sample to the oracle classification boundary. The motivation is that the smaller the margin, the larger the difficulty of a sample~\cite{JingfengZhang16}.
    \item Uncertainty-based measures. This category uses the uncertainty of a sample to measure the difficulty. Aguilar et al.~\cite{Aguilar} identify hard samples based on epistemic uncertainty and leverage the Bayesian Neural Network~\cite{Xiao} to infer it.
\end{itemize}

Varying difficulty measures have a huge impact on a difficulty-based weighting strategy. The underlying factors which influence samples' learning difficulty considered in the above measures include noise level~\cite{WenjieWang18,ThibaultCastells21}, imbalance degree~\cite{YinCui17,zhang2021distribution}, margin~\cite{JingfengZhang16}, and uncertainty~\cite{Aguilar}. However, each measure only considers a single factor or in part, and comes from heuristic inspirations but not formal certifications, hindering the application scope of the measures. It is desirable to theoretically explore a universal measure %which have a theoretical guarantee, 
capturing all of the above factors. Based on this measure, the role of difficulty-based sample weighting can be revealed more concretely. However, neither theoretical nor empirical investigations have been conducted to investigate a unified measure.

Moreover, despite the empirical success of various difficulty-based weighting methods, the process of how difficulty-based weighting positively influences the deep learning models remains unclear. Two recent studies have attempted to investigate the influence of weights in deep learning. Byrd and Lipton~\cite{JonathonByrd01} empirically studied the training of over-parameterized networks with sample weights and found that these sample weights affect deep learning by influencing the implicit bias of gradient descent-a novel topic in deep learning interpretability, focusing on why over-parameterized models is biased toward solutions that generalize well. Existing studies on this topic~\cite{DanielSoudry03,Lenaic Chizat,lyu2019gradient} reveal that the direction of the parameters (for linear predictor) and the normalized margin (for nonlinear predictor) respectively converge to those of a max-margin solution. 

Inspired by the finding of Byrd and Lipton~\cite{JonathonByrd01}, Xu et al.~\cite{xu2020understanding} dedicated to studying how the understandings for the implicit bias of gradient descent adjust to the weighted empirical risk minimization (ERM) setting. They concluded that assigning high weights to samples with small margins may accelerate optimization. In addition, they established a generalization bound for models that implement learning by using sample weights. However, they only discussed the measurement of difficulty by using one of the indicators (i.e., margin), resulting in that their conclusion is limited and inaccurate in some cases. Furthermore, their generalization bound cannot explicitly explain why hard samples are assigned with large weights in many studies. More analyses based on a universal difficulty measure are in urgent demand.

In this study, the manner of how the difficulty-based weighting affects the deep model training is deeply investigated. First, our analyses support that the generalization error of the training sample can be regarded as a universal difficulty measure for capturing all of the four factors described above. Second, based on this unified measure, we characterize the role of difficulty-based weighting on the implicit bias of gradient descent, especially for the convergence speed. Third, two new generalization bounds are constructed to demonstrate the explicit relationship between the sample weights and the generalization performance. The two bounds illuminate a new explanation for existing weighting strategies. Our study takes the first step of constructing a formal theory for difficulty-based sample weighting. In summary, our contributions are threefold.
\begin{itemize}
    \item %The generalization error is a basic concept in machine learning. 
    We theoretically prove the high relevance of the generalization error with four main factors influencing the samples' learning difficulty, further indicating that the generalization error can be used as a universal difficulty measure. 
    \item We reveal how the difficulty-based sample weighting influences the optimization dynamics and the generalization performance for deep learning. Our results indicate that assigning high weights on hard samples can not only accelerate the convergence speed but also enhance the generalization performance.
    \item We bring to light the characteristics of a good set of weights from multiple perspectives to illuminate the deep understanding of numerous weighting strategies.
\end{itemize}
\section{Preliminaries}

\subsection{Description of Symbols}

Let $\mathcal{X}$ denote the input space and $\mathcal{Y}$ a set of classes. We assume that the training and test samples are drawn \textit{i.i.d} according to some distributions $\mathcal{D}^{tr}$ and $\mathcal{D}^{te}$ over $\mathcal{X} \times \mathcal{Y}$. The training set is denoted as $T\!=\!\{\boldsymbol{x}, y\}\!=\!\{\left(\boldsymbol{x}_{i}, y_{i}\right)\}_{i=1}^{n}$ that contains $n$ training samples, where $\boldsymbol{x}_{i}$ denotes the $i$-th sample's feature, and $y_{i}$ is the associated label.  
Let $d_{i}$ and $w\left(d_{i}\right)$ be the learning difficulty and the difficulty-based weight of $\boldsymbol{x}_i$. The learning difficulty can be approximated by several values, such as loss, uncertainty and generalization error which will be explained in Section~3.

The predictor is denoted by $f\left(\boldsymbol{\theta},\boldsymbol{x}\right)$ and $\mathcal{F} =\{f\left(\boldsymbol{\theta}, \cdot\right)|\boldsymbol{\theta} \in \boldsymbol{\Theta} \subset \mathbb{R}^{\mathbbm{d}}\}$. For the sake of notation, we focus on the binary setting $y_{i} \in \{-1,1\}$ with $f\left(\boldsymbol{\theta},\boldsymbol{x}\right) \in \mathbb{R}$. The sign of the model's output $f\left(\boldsymbol{\theta},\boldsymbol{x}_{i}\right)$ is the predicted label. However, as to be clarified later, our results can be easily extended to the multi-class setting where $y_{i} \in \{1, 2, \cdots, C\}$. For multi-class setting, the softmax function is used to get the probability, and the logits are given by $\{f_{y_{j}}\left(\boldsymbol{\theta}, \boldsymbol{x}\right)\}_{j=1}^{C}$. %$\Delta f\left(\boldsymbol{x}\right)$ stands for the loss gradient of $\boldsymbol{x}$. 
Given a non-negative loss $\ell$ and a classifier $f\left(\boldsymbol{\theta}, \cdot\right)$, %we define the weighted $l$-risk of $f\left(\boldsymbol{\theta},\cdot\right)$ by $\mathcal{L} = \mathbb{E}\left[w\left(d\left(\boldsymbol{x}\right)\right)l\left(f\left(\boldsymbol{\theta},\boldsymbol{x}\right),y\right)\right]$. 
the empirical risk can be expressed as $\mathcal{L}(\boldsymbol{\theta},\boldsymbol{w})= \frac{1}{n}\sum_{i=1}^{n}w\left(d_{i}\right)\cdot \ell\left(y_{i}f\left(\boldsymbol{\theta}, \boldsymbol{x}_{i}\right)\right)$. We focus particularly on the exponential loss $\ell\left(u\right) = \exp\left(-u\right)$ and logistic loss $\ell\left(u\right) = \log\left(1+\exp\left(-u\right)\right)$. Let $\nabla l(u)$ be the loss gradient and 
$f\left(\boldsymbol{x}|T\right)$ is the trained model on $T$. The margin is denoted as $\gamma_{i}(T) = y_{i}f\left(\boldsymbol{\theta}, \boldsymbol{x}_{i}|T\right)$ for the binary setting, where it is equivalently denoted as $\gamma_{i}(T) = f_{y_{i}}\left(\boldsymbol{\theta},\boldsymbol{x}_{i}|T\right)-\max_{i \neq j}f_{y_{j}}\left(\boldsymbol{\theta},\boldsymbol{x}_{i}|T\right)$ for the multi-class setting.  

\subsection{Definition of the Generalization Error}

Bias-variance tradeoff is a basic theory for the qualitative analysis of the generalization error~\cite{TomHeskes17}. This tradeoff is initially constructed via regression and mean square error, which is given by
\begin{equation}
\begin{aligned}
Err &= \mathbbm{E}_{\boldsymbol{x},y}\mathbbm{E}_{T}[||y-f(\boldsymbol{x}|T)||_2^2]
\\& \approx \underbrace{\mathbbm{E}_{\boldsymbol{x},y}[||y-\overline{f}(\boldsymbol{x})||_2^2]}_{Bias}+\underbrace{\mathbbm{E}_{\boldsymbol{x},y}\mathbbm{E}_{T}[||f(\boldsymbol{x}|T)-\overline{f}(x)||_2^2]}_{Variance},
%\\& \approx Bias+Var
\end{aligned}
\end{equation}
\noindent where $\overline{f}\left(\boldsymbol{x}\right) = \mathbbm{E}_{T}\left[f\left(\boldsymbol{x}|T\right)\right]$. Similarly, we define the generalization error of a single sample $\boldsymbol{x}_{i}$ as
\begin{equation}
\begin{aligned}
e{\rm{rr}}_{i} &= \mathbbm{E}_{T}\left[ \ell\left(f\left( \boldsymbol{x}_{i}|T\right),y_{i}\right) \right] 
   \approx B\left(\boldsymbol{x}_{i}\right)+V\left(\boldsymbol{x}_{i}\right),
\end{aligned}
\label{sampleerror}
\end{equation}
where $B\left(\boldsymbol{x}_{i}\right)$ and $V(\boldsymbol{x}_{i})$ are the bias and variance of $\boldsymbol{x}_{i}$.
\subsection{Conditions and Definitions}
%and Other Pre-Defined Definitions
Our theoretical analyses rely on the implicit bias of gradient descent. The gradient descent process is denoted as
\begin{equation}
    \boldsymbol{\theta}_{t+1}\left(\boldsymbol{w}\right) = \boldsymbol{\theta}_t\left(\boldsymbol{w}\right) -\eta_t\nabla\mathcal{L}\left(\boldsymbol{\theta}_t\left[\boldsymbol{w}(\boldsymbol{d}\left[t\right])\right]\right),
\end{equation}
\noindent where $\eta_t$ is the learning rate which can be a constant or step-independent, $\nabla\mathcal{L}\left(\boldsymbol{\theta}_t\left[\boldsymbol{w}(\boldsymbol{d}\left[t\right])\right]\right)$ is the gradient of $\mathcal{L}$, and $\boldsymbol{w}(\boldsymbol{d}\left[t\right])$ is the difficulty-based weight of difficulty $\boldsymbol{d}$ at time $t$. The weight may be dynamic with respect to time $t$ if difficulty measures, such as loss~\cite{MPawanKumar04} and predicted probability~\cite{TsungYiLin03}, are used. %A minimal set of conditions for providing our theoretical results %for $a$-homogeneous neural networks 
%on the classification tasks are identified %. Two assumptions are considered 
%which follows existing literature~\cite{xu2020understanding}. 
To guarantee the convergence of the gradient descent, two conditions following the most recent study~\cite{xu2020understanding} are shown below. %More specific analyses are shown in the supplementary file.
\begin{itemize}
    \item The loss $\ell$ 
has an exponential tail whose definition is shown in the supplementary file. Thus, $\lim _{u \rightarrow \infty} \ell(-u)=\lim _{u \rightarrow \infty} \nabla \ell(-u)=0$. %It is easy to verify that losses including the exponential loss, log loss, and cross-entropy loss satisfy the definition.
\item The predictor $f(\boldsymbol{\theta},\boldsymbol{x})$ is $\alpha$-homogeneous such that 
$f(c \cdot \boldsymbol{\theta},\boldsymbol{x}) = c^{\alpha}f(\boldsymbol{\theta},\boldsymbol{x}), \forall c>0$.
\end{itemize}
It is easy to verify that losses including the exponential loss, log loss, and cross-entropy loss satisfy the first condition. The second condition implies that the activation functions are homogeneous such as ReLU and LeakyReLU, and bias terms are disallowed. In addition, we need certain regularities from $f(\boldsymbol{\theta}, \boldsymbol{x})$ to ensure the existence of critical points and the convergence of gradient descent:
\begin{itemize}
    \item 
    
    For $\forall \boldsymbol{x}\!\in\! \mathcal{X}$, $f(\boldsymbol{\theta},\boldsymbol{x})$ is $\beta$-smooth and $l$-Lipschitz on $\mathbbm{R}^{\mathbbm{d}}$. 
\end{itemize}
%The second assumption provides certain regularities from $f(\boldsymbol{\theta}, \boldsymbol{x})$ to ensure the existence of critical points and the convergence of gradient descent. 
The third condition is a common technical assumption whose practical implications are discussed in the supplementary file.

The generalization performance of deep learning models is measured by the generalization error of the test set $\mathcal{\hat{L}}\left(f\right)$~\cite{Goodfellow}, defined as

\begin{equation}
\begin{aligned}
&\mathcal{\hat{L}}\left(f\right)=\mathbbm{P}_{\left(\boldsymbol{x}, y\right) \sim \mathcal{D}^{te}}[{\gamma({f}\left(\boldsymbol{x}, y\right)) \leq 0}].
\end{aligned}
\end{equation}
%\vspace{-0.02in}
\subsection{Experiment Setup}
Demonstrated experiments are performed to support our theoretical analyses. For the simulated data, the linear predictor is a regular regression model, and the nonlinear predictor is a two-layer MLP with five hidden units and ReLU as the activation function. Exponential loss and standard normal initialization are utilized. CIFAR10~\cite{AlexKrizhevsky27} is experimented with, and ResNet32~\cite{KaimingHe30} is adopted as the baseline model. For the imbalanced data, the imbalance setting follows Ref.~\cite{YinCui17}. For the noisy data, uniform and flip label noises are used and the noise setting follows Ref.~\cite{JunShu28}. The models are trained with a gradient descent by using 0.1 as the learning rate. %Other values of the imbalance ratio and noise ratio are also experimented and the same conclusions are obtained.

The model uncertainty is approximated by the predictive variance of five predictions. To approximate the generalization error, we adopt the five-fold cross-validation~\cite{yang2020rethinking} to calculate the average learning error for each sample.%Margin is the average value of five runs with different seeds. 

\section{A Universal Difficulty Measure}
As previously stated, four factors pointed out by existing studies, namely, noise, imbalance, margin, and uncertainty, greatly impact the learning difficulty of samples. Nevertheless, existing measures only consider one or part of them, and their conclusions are based on heuristic inspirations and empirical observations. In this section, we theoretically prove that the generalization error of samples is a universal difficulty measure reflecting all four factors. 
%Although generalization error is a well-established concept, this is the first time that the relationship between the generalization error and the four typical factors is built with formal theories. %while existing difficulty measures only consider a single factor or in part. %The generalization error is a well established concept which is widely used in the theoretical analysis of machine learning. 
%We analyze the connection between the generalization error and the four factors influencing the samples' difficulty. 
%The exponential loss is adopted~\footnote{Our Other losses also XXX}, and the positive samples are taken as the examples in the subsequent discussion. 
All proofs are presented in the supplementary file. Without increasing the ambiguity, the generalization error of the samples is termed as error for brevity. 
\begin{figure}[t] % 纵向8行，图片靠右，宽度12.5em
    \centering
    %\vspace{-0.6cm}
    \setlength{\belowcaptionskip}{-0.02in}   %调整图片标题与下文距离
    %\hspace{-3cm}
    \includegraphics[width=0.8\linewidth]{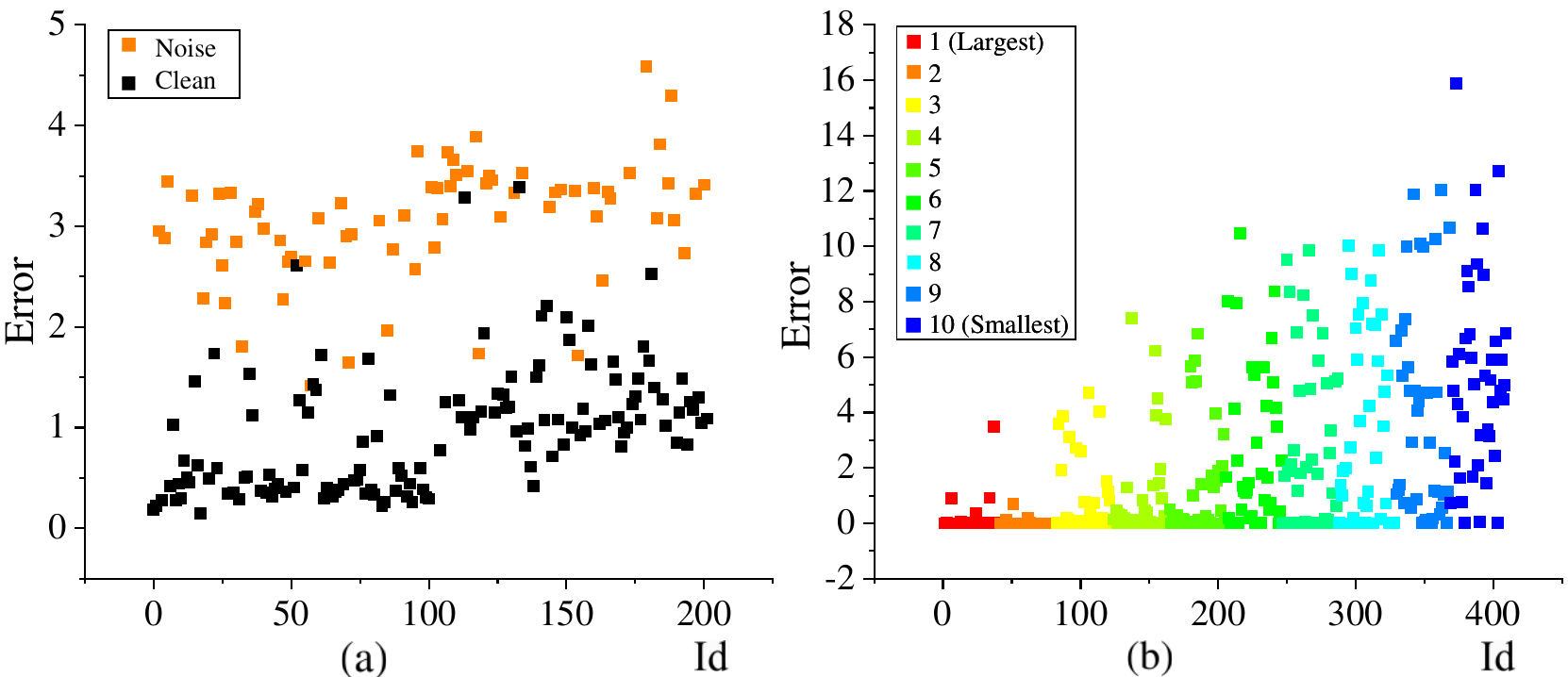} %图1b
    %\vspace{-0.15in} 
    \caption{(a) Generalization errors of clean and noisy samples on noisy data. The noise ratio is 10\% (b) Generalization errors of samples in ten categories on imbalanced data. The imbalance ratio is 10:1. CIFAR10 and ResNet32 are used. Other values of noise ratio and imbalance ratio following Ref.~\cite{JunShu28} are also experimented with and the same conclusions can be obtained.}
    \label{fig3}
    %\end{center}
\end{figure}
\subsection{Noise Factor}

Noise refers to data that is inaccurate in describing the scene. Numerous studies devoted to reducing the influence of noisy samples in the dataset on the deep learning models and these literature intuitively consider noisy samples as hard ones without formal certification~\cite{ThibaultCastells21,WonyoungShin12}. The two kinds of noise are feature noise~\cite{JelmerMWolterink13} and label noise~\cite{WonyoungShin12}. We offer two propositions to reveal the relationship between the generalization error and the noise factor.
%\subsubsection{Feature Noise}
For feature noise, we offer the following proposition:
\begin{proposition}
Let $\Delta \boldsymbol{x}_{i}$ be the perturbation of sample $(\boldsymbol{x}_{i},y_{i})$, which is extremely small in that $o(\Delta \boldsymbol{x}_{i})$ can be omitted. Let $\angle \varphi$ be the angle between the direction of $\Delta \boldsymbol{x}_{i}$ and the direction of $\mathbbm{E}_{T}[f^{\prime}\left(\boldsymbol{x}_{i}|T\right)]$. If $\mathbbm{E}_{T}[f^{\prime}\left(\boldsymbol{x}_{i}|T\right)\cdot\Delta \boldsymbol{x}_{i}]<0$ (i.e., $\angle \varphi >90^{\circ}$), then the error of the noisy sample is increased relative to the clean one. Alternatively, the direction of the perturbation $\Delta \boldsymbol{x}_{i}$ and that of $\mathbbm{E}_{T}[f^{\prime}\left(\boldsymbol{x}_{i}|T\right)]$ are contradictory. Otherwise, if $\mathbbm{E}_{T}[f^{\prime}\left(\boldsymbol{x}_{i}|T\right)\cdot\Delta \boldsymbol{x}_{i}]>0$, then $\angle \varphi < 90^{\circ}$, and the error of the noisy sample is decreased.
\end{proposition}

According to Proposition~1, feature noise can be divided into two categories, which increase or decrease the learning difficulty (generalization error) of the samples, respectively. In this paper, noise that increases the error is called the adversarial type, which is always used in the field of adversarial learning; otherwise, it is a promoted type, which refers to noise that decrease the learning difficulty of samples. Therefore, the variation of the error under feature noise is determined by the noise type. For example, as all feature noises are adversarial in adversarial learning~\cite{lowd2005adversarial}, all of the samples' errors are increased with feature noise. For label noise, we offer the following proposition:

%\subsubsection{Label Noise}
%For label noise, we offer the following proposition:
\begin{proposition}
Let $\pi$ be the label corruption rate (i.e., the probability of each label flipping to another one). Denote the probability of correct classification for the original samples as $p$. If $p>0.5$, then the errors of the noisy samples are larger than those of the clean ones.
\end{proposition}

This proposition implies that the errors of the samples with label noises are larger than those of the clean ones on the average. Specifically, if a sample is more likely to be predicted correctly, its generalization error is increased due to label noise. Let $\mathcal{L}^{*}$ be the global optimum of the generalization error of the clean dataset and $y^{\prime}$ be the corrupted label. When the noise in Proposition 2 is added, the empirical error $\mathcal{L}^\prime$ is
\begin{equation}
    \mathcal{L}^{\prime} = \left(1-\pi\right)\mathcal{L}^{*} + \pi\mathcal{L}\left(f\left(\boldsymbol{x}\right),y^{\prime}\right),
\end{equation}
where we have taken expectations over the noise. 
When $\pi \to 0$, the noise disappears, and the optimal generalization is attained. Proposition~2 is consistent with the empirical observation shown in Fig.~1(a), where the noisy samples have larger errors than the clean ones on the average. 
\subsection{Imbalance Factor}
Besides noise, imbalance is another common deviation of real world datasets. The category distribution of the samples in the training set is non-uniform. %Take the binary classification as an example, if $\mathbbm{c}_{1}$ and $\mathbbm{c}_{2}$ samples exist in the two categories, then $\mathbbm{c}_{1} \neq \mathbbm{c}_{2}$. 
Various methods solve this issue by assigning high weights on samples in tail categories which are considered to be hard ones~\cite{TsungYiLin03,YinCui17}. Nevertheless, a theoretical justification about why these samples are harder lacks.    The imbalance ratio is denoted by $c_r\!=\!{\max \{\mathbbm{c}_1, \mathbbm{c}_2, \cdots, \mathbbm{c}_C\}}\!:\!{\min \{\mathbbm{c}_1, \mathbbm{c}_2, \cdots, \mathbbm{c}_C\}}$. Then, we offer the following proposition.
\begin{proposition}
If a predictor on an imbalanced dataset ($c_r> e:1$) is an approximate Bayesian optimal classifier (as the exponential loss is an approximation for the zero-one loss), which is to minimize the total risk, then the average probability of the ground truth of the samples in the large category is greater than that of the samples in the small category.
\end{proposition}
% With Proposition~3, it is easy to obtain Proposition~4.
% \begin{proposition}
% The average generalization error $\overline{e\rm{rr}}$ of samples in the large category $\overline{e\rm{rr}}_{1}$ is larger than that of samples in the small category $\overline{e\rm{rr}}_{2}$ .
% \end{proposition}

% It indicates that there are more hard samples in the small category. 
With Proposition~3, it is easy to obtain Proposition A.1 shown in the supplementary file that the average error of samples in the small category is larger than that of the samples in the large category, indicating there are more hard samples in the small category. This proposition is verified by the experiments, as shown in Fig.~1(b). The tail categories contain more samples with larger errors. %(the imbalanced version of CIFAR10 is utilized when $c_r=10:1$)
To enhance the performance of the classification model, samples with larger errors should be assigned with higher weights, as most methods do~\cite{YinCui17}. Further experiments in Section~5 (Fig.~6) indicate that the classification performance of the small category can be improved by increasing its sample weights.
\begin{figure}[t] % 纵向8行，图片靠右，宽度12.5em
    \centering
    %\vspace{-0.6cm}
    \setlength{\belowcaptionskip}{-0.02in}   %调整图片标题与下文距离
    %\hspace{-3cm}
    \includegraphics[width=0.8\linewidth]{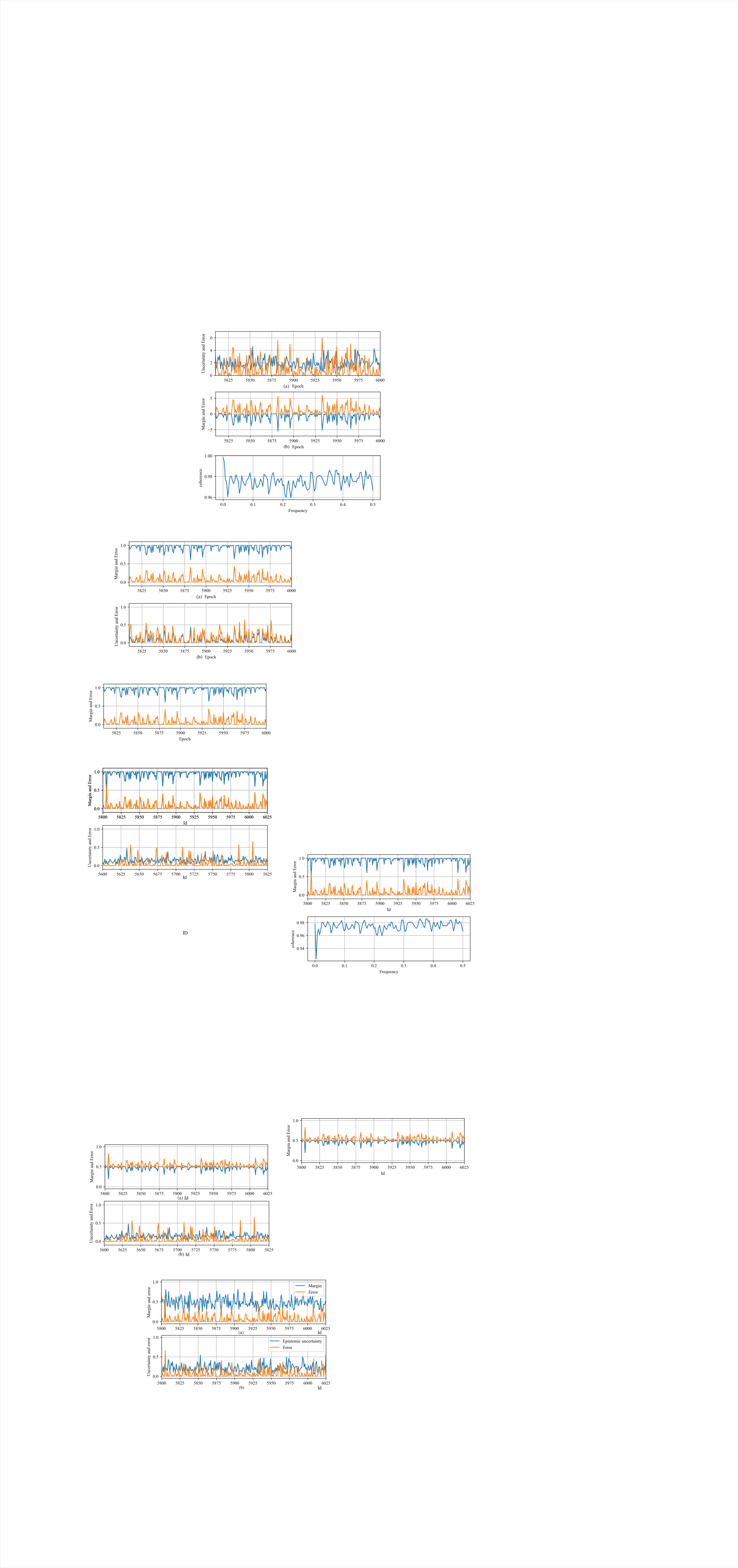} %图1b
    %\vspace{-0.15in} 
    \caption{(a) Correlation between generalization error and average margin. (b) Correlation between generalization error and epistemic uncertainty. CIFAR10 and ResNet32 are used in this experiment. All values are normalized.}
    \label{fig3}
    %\end{center}
\end{figure}

\subsection{Margin Factor}
The samples' margins measure the distances of the samples from the decision boundary. Some literature intuitively consider a small margin indicates a large learning difficulty and corresponds to a low confidence of the prediction~\cite{GamaleldinFElsayed08,JingfengZhang16}. However, a formal justification is lacking. We offer the following proposition.
\begin{proposition}
Let $\mu_i$ be the true margin of $\boldsymbol{x}_i$ corresponding to the oracle decision boundary. The condition is that the functional margins of a sample trained on random datasets obey a Gaussian distribution. In other words, for sample $\boldsymbol{x}_{i}$, its functional margin $\gamma_i$ obey a Gaussian distribution $\mathcal{N}(\mu_i, \sigma_i^2)$. For sample $\boldsymbol{x}_{j}$, $\gamma_j \sim \mathcal{N}(\mu_j, \sigma_j^2)$. %its margin $\gamma_j$ obey a Gaussian distribution.
when the margin variances of the two samples are same (i.e., $\sigma_i^2 = \sigma_j^2$), if $\mu_i\!\le\!\mu_j$, then $e{\rm{rr}}_i\!\ge\!e{\rm{rr}}_j$. Similarly, when the true margins of the two samples are the same (i.e., $\mu_i\!=\!\mu_j$), if $\sigma_i^2\!\ge\!\sigma_j^2$, then $e{\rm{rr}}_i\!\ge\!e{\rm{rr}}_j$.

\end{proposition}

Proposition~5 indicates a fact that even a sample with a large true margin, as long as the margin variance is large, it may also have a high learning difficulty. Specifically, the true margin (i.e., the mean of the functional margin distribution) of a sample and error are negatively correlated when the margin variances of the samples are equal. By contrast, the margin variance and error are positively correlated when the true margins are equal.  This illumination revises the current wisdom. The conclusion in which samples close to the oracle decision boundary are hard ones~\cite{xu2020understanding} is not completely correct. Indeed, the relation between the margin and error of sample $\boldsymbol{x}_{i}$ conforms with the following formula:
%\vspace{-0.05in}
\begin{equation}
\begin{aligned}
{e\rm{rr}}_{i} = \mathbbm{E}_{T}[e^{-\gamma_i(T)}]=e^{-\mu_i+\frac{1}{2}\sigma_i^2},
\end{aligned}
%\vspace{-0.03in}
\end{equation}
where $e\rm{rr}_{i}$, $\mu_i$, and $\sigma_i$ refer to the generalization error, the true margin, and the margin variance of sample $x_{i}$, respectively. For the two samples $\boldsymbol{x}_{i}$ and $\boldsymbol{x}_{j}$, if $\mu_i<\mu_j$ and $\sigma_i^2<\sigma_j^2$, then we cannot judge whether ${e\rm{rr}}_{i}$ is greater than ${e\rm{rr}}_{j}$. 
As shown in Fig.~2(a), the average margin and error are negatively correlated for most samples, but it is not absolute,
which accords with the above analyses. Although it is intuitive that the functional margin trained on random datasets obeys a Gaussian distribution, we evaluate it via the Z-scores of the distributions' Kurtosis and Skewness~\cite{Ghasemi} which is shown in Fig~3. More margin distribution curves and all Z-score values of the distributions are shown in the supplementary file. As all Z-scores are in $[-1.96,1.96]$, under the test level of $\alpha = 0.05$, the distribution of margin obeys the Gaussian distribution.
\begin{figure}[t] % 纵向8行，图片靠右，宽度12.5em
    \centering
    %\vspace{-0.6cm}
    \setlength{\belowcaptionskip}{-0.02in}   %调整图片标题与下文距离
    %\hspace{-3cm}
    \includegraphics[width=1\linewidth]{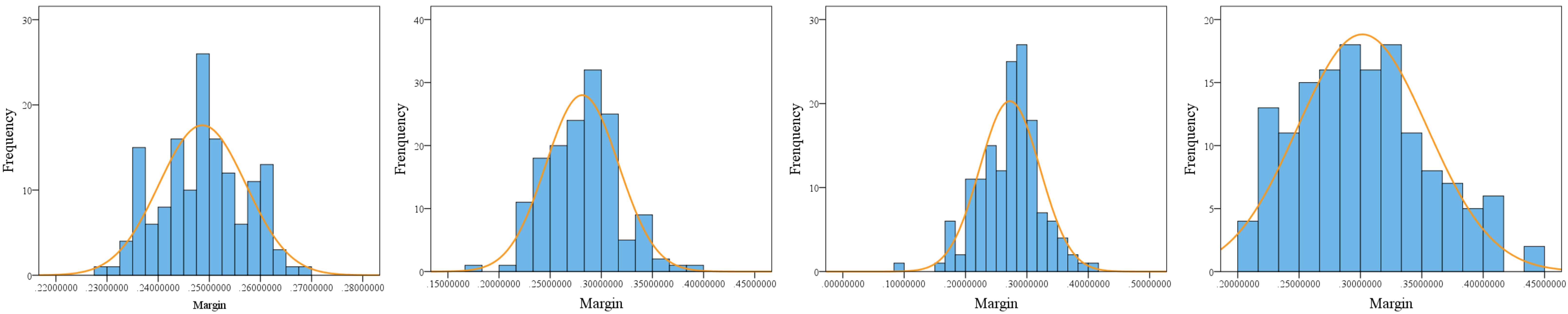} %图1b
    %\vspace{-0.15in} 
    \caption{The distributions of samples' margins.}
    \label{fig3}
    %\end{center}
\end{figure}

\subsection{Uncertainty Factor}
Uncertainties~\cite{kendall2017uncertainties} in deep learning are classified into two types. The first type is aleatoric uncertainty (data uncertainty), which is caused by the noise in the observation data. Its correlation with the error has been discussed in Section~3.1. The second type is epistemic uncertainty (model uncertainty). It is used to indicate the consistency of multiple predictions. We give the analyses of the relationship between the generalization error and epistemic uncertainty. 

Let $T$ be a training set, and let $P({\boldsymbol{\theta}}|T)$ be the distribution of the training models based on $T$. The predictive variance $Var(f(\boldsymbol{x}_{i}|\boldsymbol{\theta}_{1}),\cdots,f(\boldsymbol{x}_{i}|\boldsymbol{\theta}_K))$ plus a precision constant is a typical manner of estimating epistemic uncertainty~\cite{gal2016dropout,abdar2021review}. Take the mean square loss as an example\footnote{For other losses, other methods can be used to calculate the predictive variance~\cite{yang2020rethinking}.}, the epistemic uncertainty is
% \vspace{-0.05in}
\begin{equation}
\begin{aligned}
\widehat{\operatorname{Var}}\left[\boldsymbol{x}_{i}\right]:=&\tau^{-1}+\frac{1}{|K|} \sum\nolimits_{k}{f{(\boldsymbol{x}_{i}|\boldsymbol{\theta}_k)}^{\intercal} f{(\boldsymbol{x}_{i}|\boldsymbol{\theta}_k)}}-\mathbbm{E}[f{(\boldsymbol{x}_{i}|\boldsymbol{\theta}_k)}]^{\intercal}\mathbbm{E}[f{(\boldsymbol{x}_{i}|\boldsymbol{\theta}_k)}],
\end{aligned}
% \vspace{-0.05in}
\label{uncer6}
\end{equation}
where $\tau$ is a constant. The second term on the right side of Eq.~(\ref{uncer6}) is the second raw moment of the predictive distribution and the third term is the square of the first moment. When $K \to \infty$ and the constant term is ignored, Eq.~(\ref{uncer6}) becomes
% \vspace{-0.05in}
\begin{equation}
\begin{aligned}
\widehat{\operatorname{Var}}\left[\boldsymbol{x}_{i}\right]&:=\int_{\boldsymbol{\theta}}||f{(\boldsymbol{x}_{i}|\boldsymbol{\theta})}-\overline{f}(\boldsymbol{x}_{i})||^2_2 dP(\boldsymbol{\theta}|T).
\end{aligned}
% \vspace{-0.05in}
\
\label{uncertainty}
\end{equation}
If $P(\boldsymbol{\theta}|T)$ is approximated by the distribution of learned models on random training sets which conform to the Gaussian distribution $\mathcal{N}(T,\delta I)$, Eq.~(\ref{uncertainty}) is exactly the variance term of the error defined in Eq.~(\ref{sampleerror}) when the mean square loss is utilized. %To sum up, the model uncertainty can be represented by the variance term of the generalization error.

As the bias term in the error can capture the aleatoric uncertainty and the variance term captures the epistemic uncertainty, the overall relationship between uncertainty and error is positively correlated. Nevertheless, the relationship between epistemic uncertainty and error is not simply positively or negatively correlated.
For some samples with heavy noises, their epistemic uncertainties will be small as their predictions remain erroneous. However, their errors are large due to their large bias. This phenomenon is consistent with the experimental results shown in Fig.~2(b). Epistemic uncertainty and error are positively correlated for some samples, and the two variables are negatively correlated for other samples. %Nevertheless, epistemic uncertainty can be reflected by the generalization error. %Therefore, generalization error is a more comprehensive measure than other factors including both variance and bias. 
% \vspace{-0.02in}
\subsection{Discussion about Generalization Error}
% \vspace{-0.02in}
The commonly used difficulty measures, such as loss~\cite{MPawanKumar04} and gradient norm~\cite{CarlosSantiagoa19}, are mainly related to the bias term. Shin et al.~\cite{WonyoungShin12} emphasized that only using loss as the measurement cannot distinguish clean and noisy samples, especially for uniform label noise. There are also a few existing studies that use variance~\cite{Chang_H_S,Swayamdipta}. For instance, Agarwal et al.~\cite{agarwal2020estimating} applied the variance of gradient norms as the difficulty measure. Indeed, both the variance and bias terms should not be underestimated when measuring the samples' learning difficulty. Our theoretical analyses support that generalization error including both the two terms can capture four main factors influencing the samples' learning difficulty. Thus, the error can be leveraged as a universal measure that is more reasonable than existing measures. Existing studies generally apply the K-fold cross-validation method~\cite{yang2020rethinking} to calculate the generalization error. More efficient error calculation algorithms are supposed to be proposed %beneficial 
which will be our future work.
% \vspace{-0.02in}
\section{Role of Difficulty-Based Weighting}
% \vspace{-0.02in}
% \begin{figure*}[t] % 纵向8行，图片靠右，宽度12.5em
%     \centering
%     %\vspace{-0.6cm}
%     %\setlength{\belowcaptionskip}{-0.15cm}   %调整图片标题与下文距离
%     %\hspace{-3cm}
%     \includegraphics[width=1\linewidth]{zhengwennandujiada.pdf} %图1b
%     %\vspace{-0.15in} 
%     \caption{``Cosine distance" represents the cosine of the angle between the decision boundary (at that epoch) and the max-margin solution. (a), (b) Cosine distance and average margin of equal weights and inverse margin weights using the linear predictor. (c), (d) Cosine distance and average margin of equal weights and increasing weights of samples in the small category using the nonlinear predictor on the imbalanced data.% The training performances of the increased weights on noisy samples and other occasions are shown in Fig.~A-1.
%     } 
%     \label{fig3}
%     %\end{center}
%     %\vspace{-2pt}
% \end{figure*}
\begin{figure*}[t] % 纵向8行，图片靠右，宽度12.5em
    \centering
    %\vspace{-0.6cm}
    \setlength{\belowcaptionskip}{-0.02in}   %调整图片标题与下文距离
    %\hspace{-3cm}
    \includegraphics[width=1\linewidth]{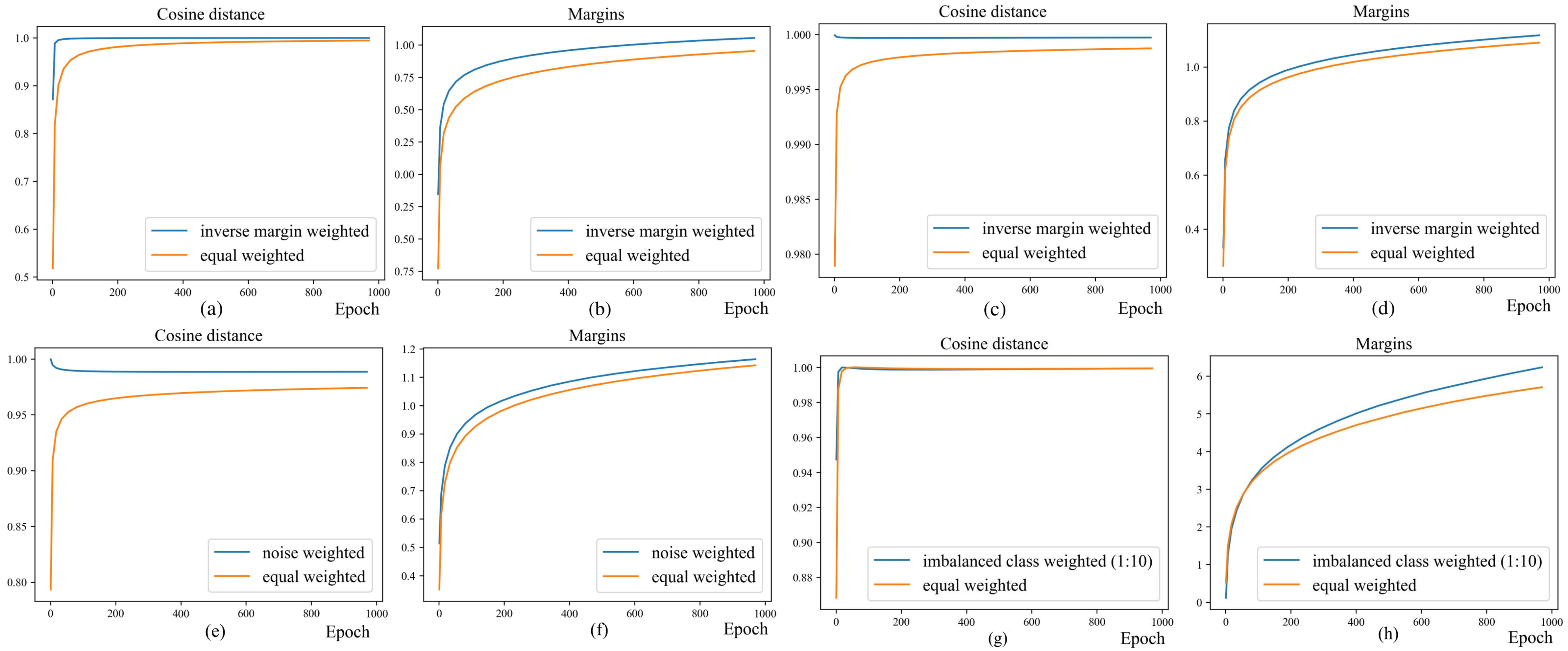} %图1b
    % \vspace{-0.15in} 
    \caption{``Cosine distance" represents the cosine of the angle between the decision boundary (at that epoch) and the max-margin solution. 
    (a), (b) Cosine distance and average margin of equal weights and inverse margin weights using the linear predictor. (c), (d) Cosine distance and average margin of equal weights and inverse margin weights using the nonlinear predictor.
    (e), (f) Cosine distance and average margin of equal weights and increasing weights of noisy samples using the nonlinear predictor on the noisy data. (g), (h) Cosine distance and average margin of equal weights and increasing weights of samples in tail categories using the linear predictor on the imbalanced data. More results are placed in the supplementary file.}
    \label{fig3}
    %\end{center}
    % \vspace{-0.02in}
\end{figure*}
This section aims to solve the second issue of explaining the difficulty-based weighting in deep learning. Based on the universal difficulty measure, the impacts of the difficulty-based weighting schemes on the optimization dynamics and the generalization performance in deep learning are investigated. Compared with the most recent conclusions \cite{xu2020understanding} established only on the margin factor, our theoretical findings, which are based on our universal measure, are more applicable and precise.
%profound.

\subsection{Effects on Optimization Dynamics}
\subsubsection{Linear Predictor}
We begin with the linear predictors allowing for a more refined analysis.  
Xu et al.~\cite{xu2020understanding} inferred an upper bound containing the term $D_{KL}(\boldsymbol{p}\|\boldsymbol{w})$, where $D_{KL}$ is the Kullback-Leibler divergence and $\boldsymbol{p}$ is the optimal dual coefficient vector. A smaller value of $D_{KL}(\boldsymbol{p}\|\boldsymbol{w})$ means that the convergence may be accelerated. Therefore, to accelerate the convergence, %reduce the upper bound which may 
they believe that the weights $\boldsymbol{w}$ should be consistent with the coefficients $\boldsymbol{p}$. Alternatively, the samples with small functional margins will have large coefficients and thus should be assigned with large weights. However, the functional margin is not the true margin that corresponds to the oracle boundary.
Therefore, their conclusion that samples close to the oracle classification boundary should be assigned with large weights~\cite{xu2020understanding} cannot be well-drawn according to their inference. We offer a more precise conclusion with the unified difficulty measure (i.e., generalization error). As before, we assume that the functional margins of a sample $\boldsymbol{x}_i$ obey a Gaussian distribution $\mathcal{N}(\mu_i,\sigma_i^2)$, where $\mu_i$ is the true margin and $\sigma_i^2$ is the margin variance of $\boldsymbol{x}_i$. We offer the following proposition:
\begin{proposition}
For two samples $\boldsymbol{x}_i$ and $\boldsymbol{x}_j$, if ${e\rm{rr}}_{i} \ge {e\rm{rr}}_{j}$, then we have:

(1) When the optimal dual coefficient ${p}_i$ of $\boldsymbol{x}_i$ on a random training set $T$ is a linear function of its functional margin $\gamma_i$ on $T$, %then,   of $x_i$ on $T$, 
if $\mu_i \le \mu_j$,  then $\mathbbm{E}_{T}[{p}_i]\ge \mathbbm{E}_{T}[{p}_j]$ (i.e., $\mathbbm{E}_{T}[{w}_i]\ge \mathbbm{E}_{T}[{w}_j]$); %Thus, samples with small true margins should be assigned large weights to speed up the convergence.

(2) When the optimal dual coefficient ${p}_i$ of $\boldsymbol{x}_i$ on a random training set $T$ is a natural exponential function of its functional margin $\gamma_i$ on $T$, $\mathbbm{E}_{T}[{p}_i]\ge \mathbbm{E}_{T}[{p}_j]$ (i.e., $\mathbbm{E}_{T}[{w}_i]\ge \mathbbm{E}_{T}[{w}_j]$) always holds. Notably, even when $\mu_{i}>\mu_j$, $\mathbbm{E}_{T}[{p}_i]>\mathbbm{E}_{T}[{p}_j]$  may still hold.
\end{proposition}

The proof is presented in the supplementary file. $\mathbbm{E}_T[p_i] > \mathbbm{E}_T[p_j]$ implies that $w_i > w_j$ holds on the average. The conclusion that samples with small true margins should be assigned with large weights may not hold on some training sets when $p_{i}$ is not a linear function of $\gamma_{i}$~\cite{DanielSoudry03}. A sample with a small true margin may have a smaller weight than a sample with a large true margin yet a large error. %For example, although the true margin of a sample is small, its margin variance may be large. 
Thus, a more general conclusion when $p_{i}$ is
not a linear function of $\gamma_{i}$ is that increasing the weights of hard samples (samples with large generalization errors) may accelerate the convergence, rather than just for samples with small margins. Other factors, including noise, imbalance, and uncertainty also affect samples' learning difficulty. Notably, the weights of the hard samples should not be excessively increased, as to be explained in the succeeding section. We reasonably increase the weights of the hard samples shown in Figs.~4 and A-3 in the supplementary file indicating that the optimization is accelerated. %The proof and the analysis for convergence direction is shown in the supplementary file.

We also prove that difficulty-based weights do not change the convergence direction to the max-margin solution shown in Theorem A.1 in the supplementary file. As shown in Fig.~3, the cosine distance and margin value are always increasing during the training procedure, indicating the direction of the asymptotic margin is the max-margin solution.% for the linear predictors on separable data, as shown in Theorem~1. The proof is placed in the supplementary.
% \begin{theorem}
% For the linear predictor with difficulty-based weights on separable data, if $\boldsymbol{\theta}^{*}$ is the $L_{2}$ max-margin vector (the solution to the hard margin SVM), 
% we have $\lim_{t\to \infty}\boldsymbol{\theta}_{t}\left[\boldsymbol{w}(\boldsymbol{d})\right] = \boldsymbol{\theta}^{*}$.
% \end{theorem}
%We see that the difficulty-based weighting does not affect the asymptotic margin for linear predictors. 

\begin{figure*}[t] % 纵向8行，图片靠右，宽度12.5em
    \centering
    %\vspace{-0.6cm}
    \setlength{\belowcaptionskip}{-0.02in}   %调整图片标题与下文距离
    %\hspace{-3cm}
    \includegraphics[width=1\linewidth]{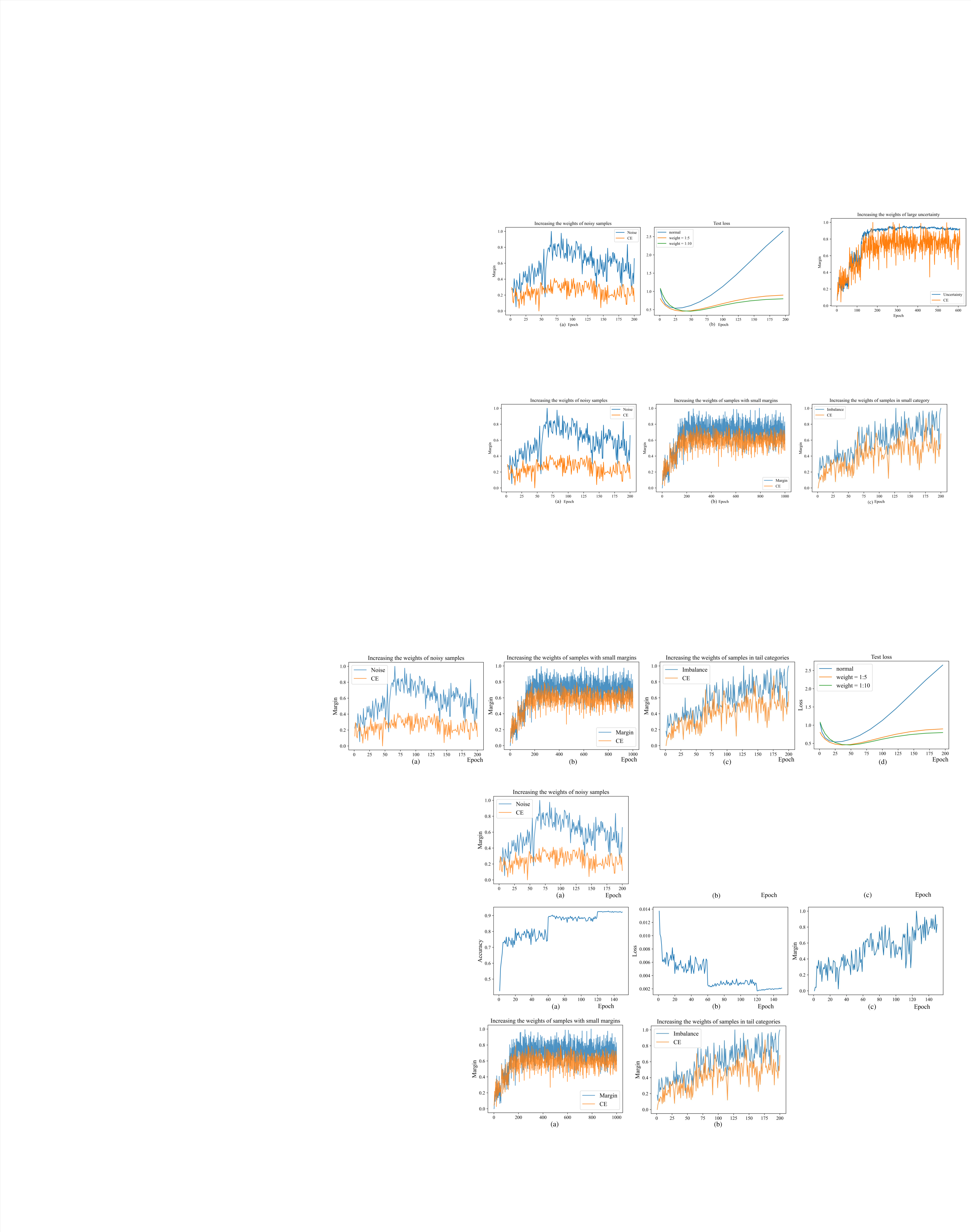} %图1b
    \caption{(a)-(c) Normalized margin of increasing the weights of noisy samples/samples with small margins/samples in tail categories. CIFAR10 data is used. Uniform label noise is adopted. The noise ratio and imbalance ratio are 10\% and 10:1. (d) Generalization error of the test set when the nonlinear model is trained with different weights on simulated imbalanced data with the imbalance ratio as 10:1. Other noise and imbalance settings are also experimented with and the same conclusions can be obtained.}
    \label{fig3}
    %\end{center}
    % \vspace{-0.02in}
\end{figure*}
\subsubsection{Nonlinear Predictor}
Analyzing the gradient dynamics of the nonlinear predictors is insurmountable. The main conclusion obtained by Xu et al.~\cite{xu2020understanding} can also be established for difficulty-based weights only if the bound of weights is larger than zero. However, their theorem has only been proven for binary cases as the employed loss is inapplicable in multi-class cases. Here, we extend the theory to the multi-class setting with a regularization $\lambda ||\boldsymbol{\theta}||^r$ on the cross-entropy loss. Let $\boldsymbol{\theta}_{\lambda}\left(\boldsymbol{w}\right)\!\in\!\arg\min \mathcal{L}_{\lambda}\left(\boldsymbol{\theta}, \boldsymbol{w}\right)$. Formally, the dynamic regime for the nonlinear predictor can be described as follows:
\begin{theorem}
Let $\boldsymbol{w} \in [b,B]^n$. Denote the optimal normalized margin as 
% \vspace{-0.05in}
\begin{equation}
    \gamma^{*}\!=\! \max_{\|\boldsymbol{\theta}(\boldsymbol{w})\|\leq 1}\min_{i} (f_{y_i}(\boldsymbol{\theta}(\boldsymbol{w}), \boldsymbol{x}_i) - \max_{j\neq i}(f_{y_{j}} (\boldsymbol{\theta}(\boldsymbol{w}), \boldsymbol{x}_i)))
\end{equation}
% \vspace{-0.05in}
Let $\overline{\boldsymbol{\theta}}_{\lambda}(\boldsymbol{w}) = {\boldsymbol{\theta}}_{\lambda}(\boldsymbol{w})/ \|{\boldsymbol{\theta}}_{\lambda}(\boldsymbol{w})\|$. 
Then, it holds that
(1) Denote the normalized margin as
\begin{equation}
\begin{aligned}
% \vspace{-0.05in}
 \gamma_{\lambda}(\boldsymbol{w})\!=\!\min_{i}(f_{y_{i}}(\overline{\boldsymbol{\theta}}_{\lambda}\left(\boldsymbol{w}\right), \boldsymbol{x}_{i})\!-\!\max_{j\neq i}f_{y_{j}}(\overline{\boldsymbol{\theta}}_{\lambda}\left(\boldsymbol{w}\right), \boldsymbol{x}_{i}))
\end{aligned}
% \vspace{-0.05in}
\end{equation}
Then, ${{\gamma}_\lambda\left(\boldsymbol{w}\right)}\!\to\!{\gamma ^*}$, as $\lambda \to 0$.

(2) There exists a $\lambda:=\lambda\left(r, a, \gamma^{*}, \boldsymbol{w}\right)$. For $\alpha\!\le\!2$, let $\boldsymbol{\theta}^{\prime}(\boldsymbol{w})$ denote a $\alpha$-approximate minimizer of $\mathcal{L}_{\lambda}$. 
Thus, $\mathcal{L}_{\lambda}\left(\boldsymbol{\theta}^{\prime}\left(\boldsymbol{w}\right)\right) \leq \alpha L_{\lambda}\left(\boldsymbol{\theta}_{\lambda}\left(\boldsymbol{w}\right)\right)$. Denote the normalized margin of $\boldsymbol{\theta}^{\prime}(\boldsymbol{w})$ by $\gamma^{\prime}\left(\boldsymbol{w}\right)$. Then,$
\gamma^{\prime}\left(\boldsymbol{w}\right) \geq \frac{\gamma^{*}}{10\alpha^{a / r}}$.
\end{theorem}

The proof is presented in the supplementary file. When $\lambda$ is
sufficiently small, the difficulty-based weighting does not affect the asymptotic margin. %Nonetheless, the following points need to be clarified. 
%\begin{itemize}
%\vspace{-0.03in}
    %\item 
According to Theorem~2, the weights do affect the convergence speed. A good property is that even though $L_{\lambda}\left(\boldsymbol{\theta}_{\lambda}\left(\boldsymbol{w}\right)\right)$ has not yet converged but close enough to its optimum, the corresponding
normalized margin has a reasonable lower bound. A good set of weights can help the deep learning model to achieve this property faster. However, the conditions in which a set of weights can accelerate the speed are not clearly illuminated. Notably, as shown in our experiments in Figs.~4 and A-3 in the supplementary file, assigning large weights for hard samples increases the convergence speed. The results on the multi-class cases (CIFAR10) indicate that assigning large weights on hard samples increases the margin, as shown in Figs.~5(a-c). 
    %\vspace{-0.03in}
However, some particular occasions of difficulty-based weights, such as SPL~\cite{MPawanKumar04}, do not satisfy the bounding condition because the lower bounds of these weights are zero instead of a positive real number. The theorem requires further revision to accommodate this situation.
    %\vspace{-0.02in}
%\end{itemize}
%Our future work will investigate a more granular solution to the convergence speed of the nonlinear predictors.

\subsection{Effects on Generalization Performance}

% \begin{figure}[t] % 纵向8行，图片靠右，宽度12.5em
%     \centering
%     %\vspace{-0.6cm}
%     %\setlength{\belowcaptionskip}{-0.15cm}   %调整图片标题与下文距离
%     %\hspace{-3cm}
%     \includegraphics[width=0.9\linewidth]{ge_cifar.pdf} %图1b
%     %\vspace{-0.1in} 
%     \caption{(a) Normalized margin of equal weights and increasing weights of noisy samples on noisy CIFAR10 data during the training procedure. 10\% uniform label noise is added. (b) Generalization error of the test set when the nonlinear model is trained with different weights on simulated imbalanced data with the imbalance ratio as 10:1. Other noise and imbalance settings are also experimented with and the same conclusions can be obtained.}
%     \label{fig3}
%     %\end{center}
%     %\vspace{-0.2in}
% \end{figure}
\begin{figure}[t] % 纵向8行，图片靠右，宽度12.5em
    \centering
    %\vspace{-0.6cm}
    \setlength{\belowcaptionskip}{-0.02in}   %调整图片标题与下文距离
    %\hspace{-3cm}
    \includegraphics[width=0.8\linewidth]{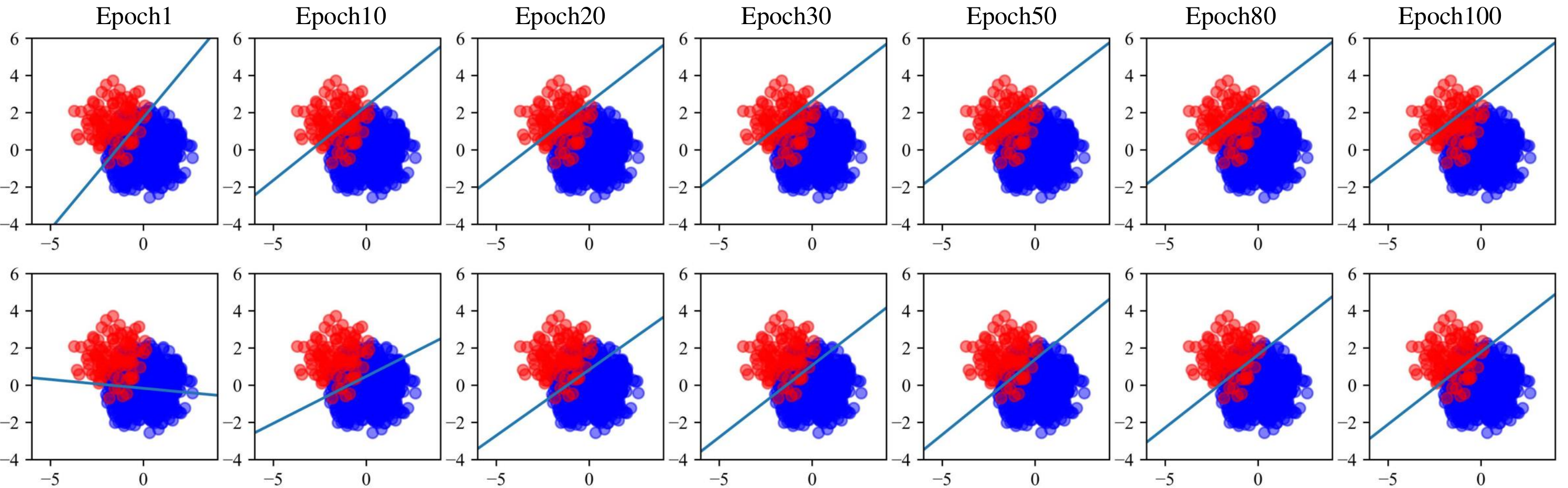} %图1b
    %\vspace{-0.1in} 
    \caption{Top: Equal weights of the two categories. Bottom: Samples in the small category are assigned with high weights, obtaining better performance for the small (red) category. The imbalance ratio is set to 10:1. The same conclusions can also be obtained for other imbalance ratios.
    }
    \label{fig3}
    %\end{center}
    % \vspace{-0.02in}
\end{figure}
Besides the role of difficulty-based weights on optimization dynamics, we are also concerned as to whether and how the difficulty-based weights affect the generalization performance. The generalization bound of Xu et al.~\cite{xu2020understanding} does not contain the sample weights, thus it cannot explicitly explain why hard samples are assigned with large weights. In addition, they assume that the source and target distributions are unequal, restricting the application of their conclusion. The two generalization bounds we propose offer good solutions to these issues. They illuminate how a weighting strategies can be designed.

Let $P_{s}$ and $P_{t}$ be the source (training) and target (testing) distributions, respectively, with the corresponding densities of $p_{s}(\cdot)$ and $p_{t}(\cdot)$. Assume that the two distributions have the same support. The training and test samples are drawn \textit{i.i.d} according to distributions $P_{s}$ and $P_{t}$, respectively. 
Learning with sample weights $\boldsymbol{w}(\boldsymbol{x})$ is equivalent to learning with a new training distribution $\widetilde{P}_{s}$. The density of the distribution of the weighted training set $\widetilde{P}_{s}$ is denoted as $\widetilde{p}_{s}(\boldsymbol{x})$ and $\widetilde{p}_{s}(\boldsymbol{x})\sim \boldsymbol{w}(\boldsymbol{x})p_s(\boldsymbol{x})$. Pearson $\chi^2$-divergence is used to measure the difference between $\widetilde{P}_{s}$ and $P_{t}$, i.e., $D_{\chi^{2}}(P_{t} \| \widetilde{P}_{s})\!=\!\int[(d \widetilde{P}_{s}/ d P_{t})^{2}\!-\!1] d \widetilde{P}_{s}$. 
We consider depth-$q$ ($q \ge 2$) networks with the activation function $\phi$. 
The binary setting is considered, in that the network computes a real value 
%\vspace{-0.02in}
\begin{equation}
\begin{aligned}
% \vspace{-0.05in}
f \left(\boldsymbol{x}\right):=\boldsymbol{W}_{q} \phi\left(\boldsymbol{W}_{q-1} \phi\left(\cdots \phi\left(\boldsymbol{W}_{1} \boldsymbol{x}\right) \cdots\right)\right),
\end{aligned}
\end{equation}
where $\phi(\cdot)$ is the element-wise activation function (e.g., ReLU). The training set contains $n$ samples. Denote the generalization error for a network $f$ as $\mathcal{{\hat{L}}}(f)$. 
The generalization performance of $f$  with weights can be described as follows.
\begin{theorem}
Suppose $\phi$ is \text{1}-Lipschitz and \text{1}-positive-homogeneous. With a probability at least of $1-\delta$, we have
\begin{equation}
% \vspace{-0.05in}
\begin{aligned}
\mathcal{{\hat{L}}}\left(f\right) &\leq \underbrace{\frac{1}{n} \sum_{i=1}^{n}{\frac{p_{t}(\boldsymbol{x}_{i})}{\widetilde{p}_{s}(\boldsymbol{x}_{i})}\mathbbm{1}(y_if(\boldsymbol{x}_i) <\gamma)}}_{I}+\underbrace{\frac{L\cdot \sqrt{D_{\chi^{2}}\left(P_{t}\| \widetilde{P}_{s}\right)+1}}{{\gamma \cdot q^{\left(q-1\right) / 2} \sqrt{n}}}}_{\left(I I\right)}+\underbrace{\epsilon(\gamma, n, \delta)}_{\left(III\right)},
\end{aligned}
% \vspace{-0.05in}
\label{fanhua}
\end{equation}
where $\epsilon(\gamma,n,\delta)=\sqrt {\frac{{\log{{\log}_2}\frac{{4L}}{\gamma }}}{n}}+\sqrt{\frac{\log \left(1/\delta \right)}{n}}$ and
$L\!:=\!\sup_{\boldsymbol{x}}\|\boldsymbol{x}\|$.
\end{theorem}
The proof is presented in the supplementary file. Compared with the findings of Xu et al.~\cite{xu2020understanding}, the bound of the generalization error is directly related to the sample weights $\boldsymbol{w}(\boldsymbol{x})$ contained in $\widetilde{p}_{s}(\boldsymbol{x})$. In view of reducing the generalization error, a natural optimization strategy can be implemented as follows: 1) an optimal weight set $\boldsymbol{w}(\boldsymbol{x})$ (in ${\widetilde{p}_{s}(x)}$) is obtained according to decreasing the right side of Eq.~(\ref{fanhua}) based on the current $f$% (increasing the weights of samples with small margins and making the test and training distributions close)
; 2)  $f$ is then optimized under the new optimal weights $\boldsymbol{w}(x)$. In the first step, the reduction of generalization error can come from two aspects. One is to increase the weights of samples with small margins. The other is to make the test and training distributions close. Disappointingly, this strategy heavily relies on the current $f$ which is unstable. Given a fixed training set, $f$ depends on random variables (denoted as $\mathcal{V}$) such as hyperparameters and initialization. To obtain a more stable weighting strategy, we further propose the following proposition.
\begin{proposition}
Suppose $\phi$ is 1-Lipschitz and 1-positive-homogeneous. With a probability of at least $1-\delta$, we have
\begin{equation}
% \vspace{-0.05in}
\begin{aligned}
\mathbbm{E}_{\mathcal{V}}[\mathcal{{\hat{L}}}\left(f_{\mathcal{V}}\right)] &\leq  \underbrace{{\frac{1}{n} \sum_{i=1}^{n}{\frac{p_{t}(\boldsymbol{x}_{i})}{\widetilde{p}_{s}(\boldsymbol{x}_{i})}\mathbbm{E}_{\mathcal{V}}[\mathbbm{1}(y_if_{\mathcal{V}}(\boldsymbol{x}_{i}) <\gamma)}}]}_{(I)}+\underbrace{\frac{L\cdot \sqrt{D_{\chi^{2}}\left(P_{t}\| \widetilde{P}_{s}\right)+1}}{{\gamma \cdot q^{\left(q-1\right) / 2} \sqrt{n}}}}_{\left(I I\right)}+ (III)
%\underbrace{\epsilon(\gamma, n, \delta)}_{\left(III\right)}
\end{aligned}
% \vspace{-0.05in}
\end{equation}
\end{proposition}
Accordingly, increasing the $\widetilde{p}_{s}(\boldsymbol{x}_{i})$ of the samples with large $\mathbbm{E}_{\mathcal{V}}[\mathbbm{1}(y_if_{\mathcal{V}}(\boldsymbol{x}_{i}) <\gamma)]$ will reduce (I). In fact, samples with larger generalization errors will have larger values of $\mathbbm{E}_{\mathcal{V}}[\mathbbm{1}(y_if_{\mathcal{V}}(\boldsymbol{x}_{i}) <\gamma)]$. The proof is placed in the supplementary file. Alternatively, increasing the weights of the hard samples %(samples with large generalization errors) 
will reduce (I). However, the weights of the hard samples cannot be increased arbitrarily as $D_{\chi^{2}}(P_{t}\| \widetilde{P}_{s})$ may be large. Therefore, a tradeoff between (I) and (II) should be attained to obtain a good set of weights. Alternatively, a good set of weights should increase the weights of hard samples while ensuring that the distributions of the training set and the test set are close.

It is worth mentioning that our two above conclusions are still insightful when $P_t\!=\!P_s$ while the conclusion of Xu et al.~\cite{xu2020understanding} assumes $P_t\!\neq\!P_s$. %However, our two above conclusions are still insightful when $P_t\!=\!P_s$. 
Apparently, even when $P_t\!=\!P_s$, assigning weights according to the samples' difficulties is still beneficial as the tradeoff between (I) and (II) still takes effect.

\section{Discussion}

Our theoretical analyses in Sections 3 and 4 provide answers to the two concerns described in Section 1.

First, the generalization error has been theoretically guaranteed as a generic difficulty measure. It is highly related to noise level, imbalance degree, margin, and uncertainty. Consequently, two directions are worth further investigating. The first direction pertains to investigating a more efficient and effective estimation method for the generalization error, enhancing its practicality. This will be our future work. As for the second direction, numerous existing and new weighting schemes can be improved or proposed using the generalization error as the difficulty measure. Our theoretical findings supplement or even correct the current understanding. For example, samples with large margins may also be hard-to-classify in some cases (e.g., with heterogeneous samples in their neighbors).

Second, the existing conclusions on convergence speed have been extended. For the linear predictors, the existing conclusion is extended by considering our difficulty measure, namely, the generalization error. For the nonlinear predictors, the conclusion is extended into the multi-class cases. Furthermore, the explicit relationship between the generalization gap and sample weights has been established. %Experiments are shown in Fig.~6(b). 
Our theorem indicates that assigning large weights on the hard samples may be more effective even when the source distribution $P_s$ and target distribution $P_t$ are equal. %(i.e., $P_s=P_t$). %the distribution of the training data $P_s$ is equal to that of the test data $P_t$. 

Our theoretical findings of the generalization bounds provide better explanations to existing weighting schemes. % but also enlighten us how a weighting strategy can be raised. %
For example, if heavy noise exists in the dataset, then the weights of the noisy samples should be decreased. 
As noisy samples are absent in the target distribution (i.e., $p_{t}(\boldsymbol{x}_{i})=0$), the weights of the noisy samples in a data set with heavy noise should be decreased to better match the source and target distributions. The experiments on the noisy data are shown in Fig.~A-5 in which decreasing the weights of noisy samples obtain the best performance. In imbalanced learning, samples in small categories have higher errors on the average. Increasing the weights of the hard samples will not only accelerate the optimization but also improve the performance on the tail categories, as shown in Figs.~5(d) and 6. These high-level intuitions justify a number of difficulty-based weighting methods. Easy-first schemes, such as Superloss~\cite{ThibaultCastells21} and Truncated loss~\cite{WenjieWang18}, perform well on noisy data. Hard-first schemes, such as G-RW~\cite{zhang2021distribution} and Focal Loss~\cite{TsungYiLin03}, are more suitable for imbalanced data.

\section{Conclusion}

This study theoretically investigates difficulty-based sample weighting. First, the generalization error is verified as a universal measure as a means of reflecting the four main factors influencing the learning difficulty of samples. Second, based on a universal difficulty measure, the role of the difficulty-based weighting strategy for deep learning is characterized in terms of convergence dynamics and the generalization bound. Theoretical findings are also presented. Increasing the weights of the hard samples may accelerate the optimization. A good set of weights should balance the tradeoff between the assigning of large weights on the hard samples (heavy training noises are absent) and keeping the test and the weighted training distributions close. These aspects enlighten the understanding and design of existing and future weighting schemes.
%\subsubsection{Acknowledgements} Please place your acknowledgments at
%the end of the paper, preceded by an unnumbered run-in heading (i.e.
%3rd-level heading).


\end{document}